\documentclass[journal,twoside,web]{ieeecolor}
\usepackage{generic}
\usepackage{cite}
\usepackage{amsmath,amssymb,amsfonts}
\usepackage{algorithmic}

\usepackage{tikz}
\usepackage{comment}
\usepackage{amsmath,amssymb} 
\usepackage{color}
\usepackage{algorithmic}
\usepackage{graphicx}
\usepackage{textcomp}
\usepackage{multirow}
\usepackage{url}

\usepackage{textcomp}
\def\BibTeX{{\rm B\kern-.05em{\sc i\kern-.025em b}\kern-.08em
    T\kern-.1667em\lower.7ex\hbox{E}\kern-.125emX}}
\begin{document}
\title{GREN: Graph-Regularized Embedding Network for Weakly-Supervised Disease Localization in X-ray Images}
\author{Baolian Qi, Gangming Zhao, Xin Wei, Changde Du, Chengwei Pan, Yizhou Yu, \IEEEmembership{Fellow, IEEE}, Jinpeng Li, \IEEEmembership{Member, IEEE}
\thanks{This work was supported in part by National Natural Science Foundation of China (62106248), Zhejiang Provincial Natural Science Foundation of China (LQ20F030013), Ningbo Public Service Technology Foundation, China (202002N3181) and Medical Scientific Research Foundation of Zhejiang Province, China (2021431314). }
\thanks{B. Qi, X. Wei, and J. Li are with Center for Pattern Recognition and Intelligent Medicine (PRIM), HwaMei Hospital, University of Chinese Academy of Sciences, NO.41 Northwest Street, Haishu District, Ningbo, Zhejiang, 315010, China. They are also with Ningbo Institute of Life and Health Industry, University of Chinese Academy of Sciences, Ningbo, Zhejiang, China.}
\thanks{G. Zhao is with Department of Computer Science, the University of Hong Kong, Hong Kong, China.}
\thanks{C. Du is with National Laboratory of Pattern Recognition, Institute of Automation, Chinese Academy of Sciences, Beijing, China.}
\thanks{C. Pan is with Institute of Artificial Intelligence, Beihang University, Beijing, China.}
\thanks{Y. Yu is with Department of Computer Science, the University of Hong Kong, Hong Kong, China.}
\thanks{B. Qi and G. Zhao contribute equally to this work. Corresponding author: Jinpeng Li (lijinpeng@ucas.ac.cn)}
}

\maketitle

\begin{abstract}
Locating diseases in chest X-ray images with few careful annotations saves large human effort. Recent works approached this task with innovative weakly-supervised algorithms such as multi-instance learning (MIL) and class activation maps (CAM), however, these methods often yield inaccurate or incomplete regions. One of the reasons is the neglection of the pathological implications hidden in the relationship across anatomical regions within each image and the relationship across images. In this paper, we argue that the cross-region and cross-image relationship, as contextual and compensating information, is vital to obtain more consistent and integral regions. To model the relationship, we propose the Graph Regularized Embedding Network (GREN), which leverages the \emph{intra-image} and \emph{inter-image} information to locate diseases on chest X-ray images. GREN uses a pre-trained U-Net to segment the lung lobes, and then models the intra-image relationship between the lung lobes using an intra-image graph to compare different regions. Meanwhile, the relationship between in-batch images is modeled by an inter-image graph to compare multiple images. This process mimics the training and decision-making process of a radiologist: comparing multiple regions and images for diagnosis. In order for the deep embedding layers of the neural network to retain structural information (important in the localization task), we use the Hash coding and Hamming distance to compute the graphs, which are used as regularizers to facilitate training. By means of this, our approach achieves the state-of-the-art result on NIH chest X-ray dataset for weakly-supervised disease localization. Our codes\footnote{https://github.com/qibaolian/GREN} are accessible online. 
\end{abstract}


\section{Introduction}
\label{sec:introduction}
\IEEEPARstart{T}{he} automatic disease localization on chest X-ray images has become an increasingly important technique to support clinical diagnosis and treatment. The past decade has witnessed the application of convolutional neural networks (CNNs) in medical image analysis, including disease classification~\cite{yan2018weakly},~\cite{irvin2019chexpert},~\cite{avilesrivero2020graphxnet},~\cite{JBHIClassification9721639}, segmentation~\cite{chen2018semantic},~\cite{gaal2020attention},~\cite{viniavskyi2020weakly},~\cite{JBHISegmentation9699053}, detection~\cite{zhu2018deepem},~\cite{cai2018iterative}, and report generation~\cite{jing2017automatic},~\cite{li2018hybrid},~\cite{wang2018tienet}. CNNs require a large amount of finely annotated training data to locate diseases. However, the annotation is tedious and laborious for radiologists. As a result, the X-ray datasets often only have the \emph{image-level labels} and few \emph{location labels}, which leads to inferior localization results. Developing localization models with coarse-grained (image-level) labels and few location annotations is an urgent problem, which 
arouses a weakly-supervised localization task~\cite{10.1093/nsr/nwx106}.

Existing weakly-supervised localization algorithms are mainly based on multi-instance learning (MIL)~\cite{babenko2008multiple},~\cite{li2018thoracic} and class activation maps (CAM)~\cite{Grad-CAM},~\cite{wang2017chestx},~\cite{tang2018attention}. However, these methods often yield inaccurate or incomplete regions. For example, the CAM-based method only activates the most discriminative region and generates incomplete targets. To overcome these problems, beyond mere algorithmic innovations~\cite{sedai2018deep},~\cite{cai2018iterative},~\cite{hermoza2020region},~\cite{tam2020weakly}, we argue that considering the complementary information between anatomical regions and multiple images can lead to a better localization result. This insight is inspired by the domain knowledge in the medical field: radiologists are trained to read a large number of X-ray images and analyse them by recognizing and comparing the differences (e.g., shapes, textures, contrast, etc.). Radiologists make decisions based on what they have learned and observed though comparison.

Recently, several attention-based works~\cite{liu2019align},~\cite{zhou2021contrast} conducted disease localization by comparing abnormal and normal instances. However, they involved no direct supervision (e.g., the position label) to conduct error feedback, and it was difficult to interpret what the models have learned.
To overcome these insufficiencies, we proposed the cross chest graph (CCG)~\cite{DBLP:conf/mm/Zhao21} to regularize the model instead of the attention operations~\cite{liu2019align},~\cite{zhou2021contrast}. CCG considered the inter-image relationship and neglected the intra-image relationship. More importantly, CCG obtained regions via clustering, and the self-organized regions lack pathological explanations. Based on but quite different from CCG, the proposed Graph Regularized Embedding Network (GREN) (Figure \ref{fig1}) considers both inter-image and intra-image relationships, and explicitly models the anatomical structures. We calculate the relationship between the two lung lobes because most chest diseases such as pneumonia, infiltration and consolidation, rarely appear symmetrically in both sides~\cite{zhou2021contrast}, and cardiomegaly usually makes left lung region smaller on X-ray image. Thus, the features of these abnormal regions can be identified via comparing the lung lobes. Our main contributions are:

\begin{figure}[!t]
\centerline{\includegraphics[width=0.98\columnwidth]{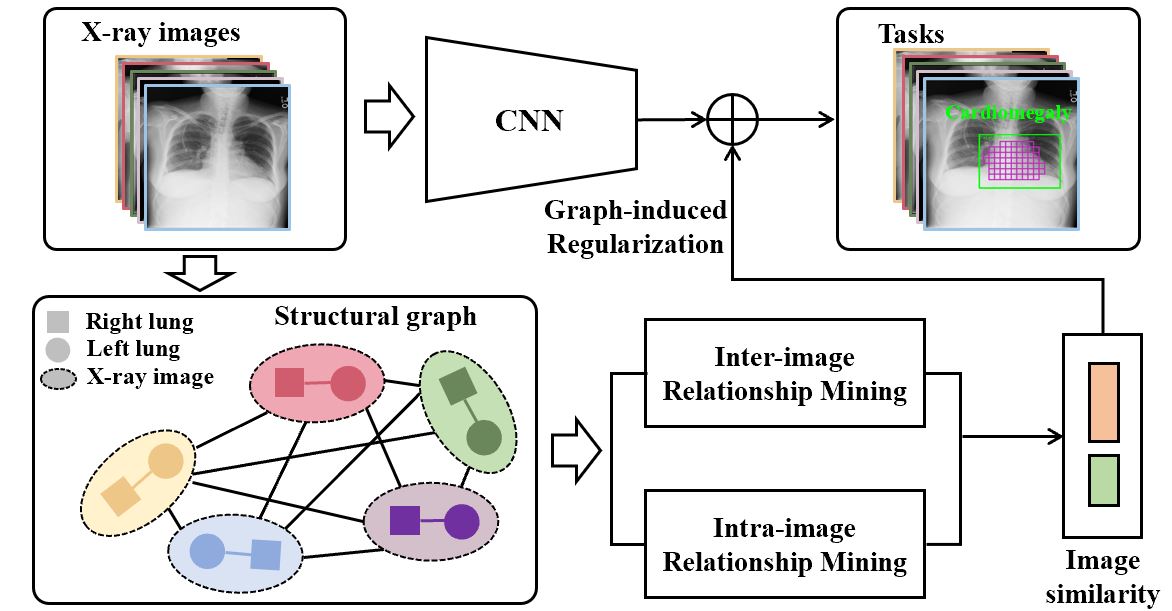}}
\caption{The general design of GREN. The inter-image and intra-image relations modeled by structural graphs are used to regularize CNNs to obtain accurate and complete regions. GREN mimics a radiologist by leveraging cross-image and cross-region information during training and decision-making.}
\label{fig1}
\end{figure}

\begin{itemize}
\item We propose GREN, the first method exploring relational graphs across images and regions to regularize training in weakly-supervised localization. Compared with the attention-based methods, GREN avoids the normal control sampling problem. Compared with Graph-Rise~\cite{juan2019graphrise}, GREN requires no additional manual annotations. As the core elements, we present the intra-image and inter-image knowledge learning modules to model the complementary information.

\item  We achieve the state-of-the-art result for disease localization on NIH chest X-ray dataset. Ablation experiments demonstrate the effectiveness of each component. We make the code publicly available for subsequent studies to reproduce our results.
\end{itemize}

\section{Related Work}

Deep learning has been widely applied in X-ray images analysis, including disease classification~\cite{wang2017chestx},~\cite{Tang},~\cite{hermoza2020region},~\cite{zhao2021diagnose} and detection~\cite{li2018thoracic},~\cite{liu2019align},~\cite{tam2020weakly}. However, locating diseases in chest X-ray images with few careful annotations remains a challenging problem. Wang et al.~\cite{wang2017chestx} contributed a chest X-ray dataset and localized the disease with a multi-label disease classification model according to the heatmaps. Li et al.~\cite{li2018thoracic} proposed to jointly model the disease identification and localization with multi-instance loss and binary cross-entropy loss. Tang et al.~\cite{Tang} proposed an iterative attention-guided refinement framework to improve the classification and weakly-supervised localization performance via CAM~\cite{Grad-CAM}. Sedai et al.~\cite{sedai2018deep} proposed a weakly supervised method based on class-aware multiscale convolutional feature to localize chest pathologies. Cai et al.~\cite{cai2018iterative} proposed an attention mining strategy to improve the sensitivity and saliency of model to disease patterns. All these methods overlooked the image-to-image and region-to-region relationship during modelling.

For the last three years, some works have attempted to embed image relations into deep learning. Zhao et al.~\cite{zhao2021contralaterally} modeled the contralateral context information using a spatial transformer network to enhance disease representations. Lian et al.~\cite{lian2021structure} leveraged the constant structure and disease relations using a structure-aware relation extraction network. These methods utilized the region-to-region relationship in individual image and overlooked the image-to-image relationship in modelling. There are also some works using inter-image relationship for weakly-supervised medical image analysis. Liu et al.~\cite{liu2019align} utilized the contrast-induced attention acquired on paired images between healthy and unhealthy samples to provide more information for localization. Zhou et al.~\cite{zhou2021contrast} exploited two contrastive abnormal attention models to improve the performance of thoracic multi-disease recognition. They designed a left-right lung contrastive network to learn intra-attentive abnormal features, and an inter-contrastive abnormal attention model to compare healthy samples with multiple unhealthy samples to compute the abnormal attention map. However, the uncertainty of the normal control may lead to the uncertainty of performance. Regularizing the models, as another strategy, involves no such sampling problem. Juan et al.~\cite{juan2019graphrise} proposed a neural graph learning framework leveraging graph structures to regularize training, where the similarity between images is used as edges. Graph-Rise~\cite{juan2019graphrise} used the distance graph between images for regularization. CCG~\cite{DBLP:conf/mm/Zhao21} constructed graph nodes using clustered regions and lacked interpretability. The proposed GREN, in contrast, is free from additional manual annotation and the regions has explicit anatomical definitions.

\begin{figure*}[!t]
\centerline{\includegraphics[width=0.95\textwidth]{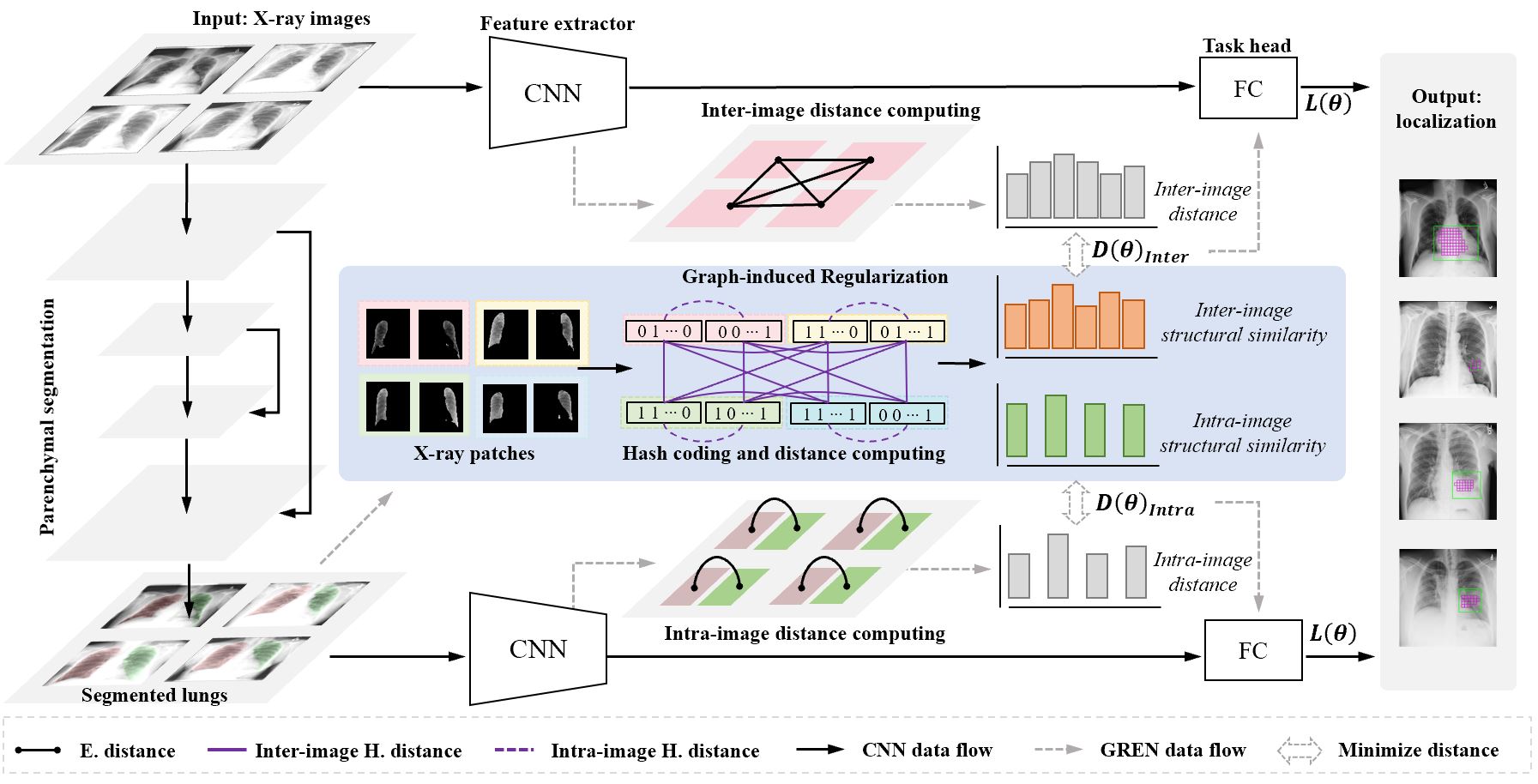}} 
\caption{Flowchart of the proposed GREN. The main part conducts weakly-supervised localization using a CNN-based feature extractor and a task head optimized via MIL. The graph-induced regularization module (in blue background) models the inter-image and intra-image similarities, which are used to regularize the representations of the main task for better structural and contextual awareness. The CNNs (FCs) refer to the same model.}
\label{fig2}
\end{figure*}

\section{Method}

The architecture of GREN is presented at Figure \ref{fig2}. The input images $X = \{x_1, x_2, ... , x_n\}$ are chest X-ray images in a training batch.   
They go through a parenchymal segmentation network~\cite{lungs-segmentation}, and the left lung region and right lung region of each image are thus obtained, which is $X^{'}=\{x_1^l, x_1^r, x_2^l, x_2^r, ... , x_n^l,x_n^r\}$. Then, the input images $X$ and the images of lung regions $X^{'}$ are fed to the feature extractor to yield feature maps $F_{Inter} = \{f_1, f_2, ... , f_n\}$ and $F_{Intra} = \{f_1^l,f_1^r, f_2^l, f_2^r, ... , f_n^l, f_n^r\}$. The distance of the feature map $F_{Intra}$ and the feature map $F_{Inter}$ is computed to obtain the intra-image distance and inter-image distance, respectively. At the same time, the Hash coding with $M$ digits is used to represent each patch. We calculate the Hamming distances between paired patches, resulting into the intra-image and inter-image similarity. Finally, the intra-image similarity and inter-image similarity are used to minimize the intra-image distance and the inter-image distance, respectively. A task head is added upon the embedding space to complete the weakly-supervised localization task. Please note that the feature extractor and the task head share weights. GREN aims to utilize the structural relationship of images and formulate a more informative embedding space to improve the localization performance. 

\subsection{Backbone}

The ResNet-50 pre-trained on the ImageNet dataset is used as the feature extractor of GREN. The input images are $X = \{x_{1}, x_{2},... , x_{n}\},x_{i}\in \mathbb{R}^{512\times 512}$. After removing the final classification and global pooling layer of ResNet-50, the feature map $F = \{f_{1}, f_{2},... , f_{n}\},f_{i}\in \mathbb{R}^{2048\times 16\times 16}$ is extracted, which is 32 times down-sampled compared with the input. Meanwhile, the input image is divided into $P\times P$ patch grids, and for each grid, the existent probability of disease is predicted by the network. Note that $P$ is an adjustable hyper-parameter ($P$=16 in our experiments). Then, we pass $F$ through two $1\times 1$ convolutional layers and a sigmoid layer to obtain the final predictions of $P \times P$ with $K$ channels, where $K$ is the number of possible disease types. Finally, we compute losses and make predictions in each channel for the corresponding class, following the paradigm used in~\cite{liu2019align} and~\cite{li2018thoracic}.  

For images with \emph{box-level} annotations, if the grid in the feature map is covered partially by the ground truth box, we assign label 1 to the grid. Otherwise, we assign 0 to it. We use the binary cross-entropy (BCE) loss for each grid:

\begin{equation}
L^k_i(\theta)_1= \sum_j -y_{ij}^k \log(p_{ij}^k)
 - \sum_j (1-y_{ij}^k) \log(1-p_{ij}^k),
\end{equation}
where $k$, $i$, and $j$ are the indexes of class, sample and grid, respectively. $y^k_{ij}$ denotes the target label of the $j$th grid in the $i$th image for the class $k$, and $p^k_{ij}$ denotes the predicted probability of the $j$th grid in the $i$th image for the class $k$.

For images with only \emph{disease-level} annotations, we use the MIL loss:

\begin{equation}
\begin{split}
L^k_i(\theta)_2 = -&y^k_i\log(1-\prod_{j} (1-p^k_{ij}))\\
-&(1-y^k_i)\log(\prod_{j} (1-p^k_{ij})),
\end{split}
\end{equation}
where $y^k_i$ denotes the target label of the image. For the backbone network, the loss $L(\theta)$ is formulated as
\begin{equation}
\begin{split}
L(\theta) = \sum_i \sum_k \beta \lambda^k_i  L^k_i(\theta)_1
 + (1-\lambda^k_i) L^k_i(\theta)_2,
\end{split}
\end{equation}
where $\theta$ denotes the network parameters, $\lambda^k_i \in {0,1}$ denotes if the $k_{th}$ class in the $i_{th}$ sample has the box annotation,
and $\beta$ is the balance weight of the two losses and is set to 4.


\subsection{Graph Regularization}

Although existing works indeed paved the way to locate and identify diseases in chest X-ray images with limited box-level labels, there are still abundant information about the inter-image and intra-image structural relationship (similarity) that can be exploited to compensate for the lack of labels. These structural similarity can be naturally represented as a graph, where the graph nodes denote the images and the edges denote their relationships. To model the structural similarities, we train the network using the graph regularization. Namely, \emph{the more similar the images are, the closer their representations are in the embedding space.} To achieve this goal, we exploit two graph regularization terms to encourage similar images (within batches) to move closer in the embedding space.

In an X-ray image, the areas containing the common lesions are the left and right lung regions, and the clavicle area and the blank area outside the body contain rare information. Therefore, the left and right lung regions of all samples are extracted by the segmentation algorithm~\cite{lungs-segmentation}. The similarity is then calculated to measure the structural relationships of lung regions. We opt to computational efficiency and choose the Hamming distance and cosine distance for CNNs during computation.

Hash coding~\cite{2000Robust} is an algorithm that produces a snippet or fingerprint of images, which analyses the image structure and shows whether two images look nearly identical. We construct the relation graph based on Hash coding for two reasons. First, Hash coding has a higher computation and storage efficiency, which can fast calculate the Hamming distance to obtain the similarity of two images without additional manual annotation~\cite{10.5555/3294996.3295009}. Second, Hash coding preserves structural information of X-ray images, which is ideal for the localization task. Since it generates a low frequency signature in the spectrum space, the low frequency signature of the monochromatic X-ray image can be viewed as structural information. The insight has been verified in~\cite{li2018thoracic} and~\cite{liu2019align}. Thus, we use the Hash coding and Hamming distance to measure the similarity of X-ray images. Specifically, the perceptual image Hash is designed not to change much when an image undergoes minor modifications such as compression, color-correction, and brightness change~\cite{samanta2021analysis}.

The computation has three steps, including image pre-processing, perceptual feature extraction, and quantization or compression to generate the Hash string. First, an image is pre-processed by resizing, color transformation and normalization. This step reduces the size of the data, and thus reducing the processing time. Second, the discrete cosine transform (DCT) is used to make perception feature extraction of an image invariable to content preserving manipulation. Third, numerical values that represent features of an image are quantized, generating a fixed sized Hash which is a compact and somewhat unique representation of an image~\cite{samanta2021analysis}. Finally, the Hamming distance is used to quantify the similarity of two Hash codings.
Additionally, the cosine similarity is a measurement that quantifies the similarity between two or more vectors, which is the cosine of the angle between vectors, and it is most commonly used in high-dimensional features of CNNs~\cite{spacesenwiki}. We use the cosine distance as another similarity algorithm to measure the similarity of the CNN features. 
\subsubsection{Intra-image Knowledge Learning}
The left and right lung regions in chest X-ray image show some symmetric structures, and the lesions of most diseases rarely appear symmetrically in both sides, such as pneumonia, infiltration and consolidation~\cite{zhou2021contrast}. It is very useful to compare and distinguish the differences (e.g., textures and shadows) between the two lung regions for radiologists.

The structural relationship of X-ray regions is modeled as a graph $G_{Intra} = (N,E)$, where the nodes $N$ denote the two regions of left lung and right lung. The edges $e_i^{lr}\in E$ denote the similarity of the graph nodes $N$, 
\begin{equation}
e_i^{lr} = \frac{1}{D_{e}} H(h_i^{l}, h_i^{r}),
\end{equation}
where $h_i^{l}$ and $h_i^{r}$ are the representations of the Hash coding of the left lung regions $x_i^{l}$ and the right lung regions $x_i^{r}$ in $i_{th}$ image, respectively. It is obtained by using perceptual hashing algorithm. $H(h_i^{l}, h_i^{r})$ calculates the Hamming distance of $h_i^{l}$ and $h_i^{r}$. $D_{e}$ is a normalized parameter, which makes $e_i^{lr}\in (0,1)$. The value of it is the product of the length of the Hash coding  ($D_{e}$ = 64 in our experiments).
The graph regularization term $D(\theta)_{Intra}$ is obtained, 
\begin{equation}
D(\theta)_{Intra} = -\sum_i e_i^{lr}d(f_i^{l}, f_i^{r}),
\end{equation}
where $e_i^{lr}$ denotes the similarity of the left lung region $x_i^{l}$ and the right lung region $x_i^{r}$, $d(*)$ is the distance metric function, which is the Euclidean distance, $f_i^{l}$ and $f_i^{r}$ mean the feature map of the left lung region $x_i^{l}$ and the right lung $x_i^{r}$, respectively. Please note that the $f_i^{l}$ and $f_i^{r}$ are obtained by taking the masks of the left lung region $x_i^{l}$ and the right lung region $x_i^{r}$ in the feature map $f_i$. 

\subsubsection{Inter-image Knowledge Learning}

There are phenomena of “different X-ray images of the similar diseases" and “different diseases of the similar X-ray images", so it is difficult for the deep learning models to locate and classify diseases with limited box-level labels. Therefore, we propose a inter-image knowledge learning module to dig out the structural differences between different X-ray images, which provides a large difference for the similar diseases and similar X-ray images. 

A pair of images $x_{u}$ and $x_{v}$ are randomly selected from the input images $X$. The structural relationship information of different X-ray images is modeled to get the graph $G_{Inter} = (N,E)$, where the graph nodes $N$ denote the images in a mini-batch of the network ($N = 4$ in our experiments), the edges $e_{uv}\in E$ denote the similarity of graph nodes $N$. Since there are left and right lung regions in an image, the Hamming distance of two lung regions in all images should be calculated,
\begin{equation}
\begin{split}
e_{uv} = \frac{1}{2D_{e}}(H(h_{u}^l, h_{v}^l)+H(h_{u}^r, h_{v}^r)),
\end{split}
\end{equation}
where $h_{u}^l$, $h_{u}^r$, $h_{v}^l$ and $h_{v}^r$ are the Hash codings of left and right lung regions of the image $x_{u}$ and the image $x_{v}$, respectively. $H(*)$ is the Hamming distance calculation. Note that the Hash coding generation is training-free. $D_{e}$ is a normalization parameter, which denotes the product of the length of the Hash coding  ($D_{e}$ = 64 in our experiments).
The graph regularization term $D(\theta)_{Inter}$ is written as
\begin{equation}
D(\theta)_{Inter} = -\sum_{u,v} e_{uv}d(f_{u}, f_{v}),
\end{equation}
where $e_{uv}$ denotes the similarity of $x_{u}$ and $x_{v}$, and $d(*)$ is the Euclidean distance, and $f_{u}$ and $f_{v}$ mean the feature maps of image $x_{u}$ and image $x_{v}$, respectively.

The final objective $Q(\theta)$ is the summation of the loss $L(\theta)$ of the localization network and the graph regularization terms $D(\theta)_{Intra}$ and $D(\theta)_{Inter}$,
\begin{equation}
Q(\theta) = L(\theta) + \lambda_1 D(\theta)_{Intra} + \lambda_2 D(\theta)_{Inter},
\end{equation}
where the hyper-parameters $\lambda_1$ and $\lambda_2$ control the balance between the three losses and are set to 0.11 and 0.15 by searching from 0 to 1 with a step of 0.05 and 0.01 for coarse-to-fine searching. For example, it is assumed that the hyper-parameter in coarse searching is 0.10 (with the step of 0.05), and the interval in fine searching is [0.96, 0.99] and [0.11, 0.14] (with the step of 0.01). When $\lambda_1$ = 0 and $\lambda_2$ = 0, the network reduces to a localization model without the graph regularization.
\begin{table*}[t]
\center
\caption{Performance comparison of disease localization using 50\% unannotated images and 80\% annotated images. Note that the data partition settings are inherited from existing studies for fair comparisons. For each column, the bold or red values denote the best results.}
\resizebox{0.98\textwidth}{!}{
\begin{tabular}{l c c c c c c c c c c}
\hline
T (IoU) & Models & Atelectasis& Cardiomegaly & Effusion & Infiltration & Mass & Nodule & Pneumonia & Pneumothorax & Mean \\
\hline
\multirow{8}{*}{0.5}& Wang et al.~\cite{wang2017chestx} & 0.05 & 0.18 & 0.11 & 0.07 & 0.01 & 0.01 & 0.03 & 0.03 & 0.06 \\
    & Li et al~\cite{li2018thoracic} & 0.14 & 0.84 & 0.22 & 0.30 & 0.22 & 0.07 & 0.17 & 0.19 & 0.27 \\
    & Liu et al.~\cite{liu2019align} & 0.32 & 0.78 & 0.40 & 0.61 & 0.33 & 0.05 & 0.37 & 0.23 & 0.39 \\
    & Zhou et al.~\cite{zhou2021contrast} & \textbf{0.39} & 0.86 & 0.46 & 0.65 & 0.39 & 0.13 & 0.43 & 0.27 & 0.45 \\
	 & Zhao et al.~\cite{DBLP:conf/mm/Zhao21} & 0.27 & 0.86 & 0.48 & 0.72 & \textbf{0.53} & 0.14 & \textbf{0.58} & 0.35 & 0.49 \\
    \cline{2-11}  
    & Baseline (\textbf{B}) & 0.12 & 0.79 & 0.32 & 0.36 & 0.26 & 0.07 & 0.13 & 0.20 & 0.28 \\ 
    & \textbf{B}+Intra & 0.37 & 0.86 & 0.48 & 0.80 & 0.37 & 0.14 & \textbf{0.58} & 0.30 & 0.49 \\
    & \textbf{B}+Inter & 0.32 & \textbf{0.89} & \textbf{0.52} & 0.80 & 0.42 & 0.21 & 0.50 & 0.35 & 0.50 \\
    & GREN\ (\textbf{B}+Intra+Inter) & 0.37 & 0.86 & \textbf{0.52} & \textbf{0.84} & 0.42 & \textbf{0.29} & 0.54 & \textbf{0.45} & {\color{red}\textbf{0.54}}  \\
\hline
\multirow{8}{*}{0.7} & Wang et al.~\cite{wang2017chestx} & 0.01 & 0.03 & 0.02 & 0.00 & 0.00 & 0.00 & 0.01 & 0.02 & 0.01 \\
    & Li et al.~\cite{li2018thoracic} & 0.04 & 0.52 & 0.07 & 0.09 & 0.11 & 0.01 & 0.05 & 0.05 & 0.12 \\
    & Liu et al.\cite{liu2019align} & 0.18 & 0.70 & 0.28 & 0.41 & 0.27 & 0.04 & 0.25 & 0.18 & 0.29 \\
    & Zhou et al.~\cite{zhou2021contrast} & 0.24 & 0.75 & 0.33 & 0.45 & 0.33 & 0.09 & 0.35 & 0.23 & 0.35 \\
	& Zhao et al.~\cite{DBLP:conf/mm/Zhao21} & 0.20 & \textbf{0.86} & \textbf{0.48} & 0.68 & 0.32 & 0.14 & 0.54 & 0.30 & 0.44 \\
    \cline{2-11}  
    & Baseline (\textbf{B}) & 0.05 & 0.71 & 0.16 & 0.04 & 0.16 & 0.00 & 0.08 & 0.15 & 0.17 \\ 
    & \textbf{B}+Intra &0.27 & \textbf{0.86} & 0.44 & \textbf{0.72} & 0.32 & 0.14 & \textbf{0.58} & 0.20 & 0.44\\
    & \textbf{B}+Inter & 0.29 & \textbf{0.86} & \textbf{0.48} & 0.68 & \textbf{0.37} & \textbf{0.21} & 0.46 & \textbf{0.35} & {\color{red}\textbf{0.46}}\\
    & GREN\ (\textbf{B}+Intra+Inter) & \textbf{0.34} & \textbf{0.86} & \textbf{0.48} & 0.60 & \textbf{0.37} & \textbf{0.21} & 0.46 & \textbf{0.35} & {\color{red}\textbf{0.46}} \\
\hline
\end{tabular}}
\label{table 1}
\end{table*}
\subsubsection{CNN Features Instead of Hash Coding}

To calculate the similarity between regions, it is most intuitive to use CNN features. To investigate on this, we replace the similarity computing with CNN features to compare with the Hash coding with Hamming distance. The feature map $F_{Intra} = \{f_1^l,f_1^r, f_2^l, f_2^r, ... , f_n^l, f_n^r\}$ is extracted using the same method as described previously. The similarities between the feature maps are calculated according to 
\begin{equation}
s_{uv} = \frac{1}{2}(d_{cos}(f_u^l, f_v^l)+ d_{cos}(f_u^r, f_v^r)), s^{lr}_{i} = d_{cos}(f^l_i, f^r_i),
\end{equation}
where $f_u^{l(r)}$ and $f_v^{l(r)}$ are the feature maps of the left (right) regions of the image $x_{u}$ and image $x_{v}$. $s_{uv}$ is the similarity of feature maps $f_u$ and $f_v$. $s^{lr}_{i}$ is the similarity of the right regions and left regions in $i_{th}$ feature maps. $d_{cos}(*)$ is the cosine distance of the feature maps. The graph regularization term $D(\theta)_{Inter}^s$ and $D(\theta)_{Intra}^s$ are written as
\begin{equation}
\begin{aligned}
& D(\theta)^{s}_{Inter} = -\sum_{u,v}s_{uv}d(f_{u}, f_{v}),\\
& D(\theta)^{s}_{Intra} = -\sum_{i}s^{lr}_{i}d(f_{i}^l, f_{i}^r),
\end{aligned}
\end{equation}
where $d(*)$ is the Euclidean distance metric. $f_{u}$ and $f_{v}$ mean the feature maps of image $x_{u}$ and image $x_{v}$, respectively.
The final objective $Q(\theta)^s$ is the summation of the localization loss $L(\theta)$ and the graph regularization loss $D(\theta)_{Inter}^s$ and $D(\theta)_{Intra}^s$,
\begin{equation}
Q(\theta)^s = L(\theta) + \lambda_3 D(\theta)^{s}_{Inter} + \lambda_4 D(\theta)^{s}_{Intra},
\end{equation}
where the hyper-parameter $\lambda_3$ and $\lambda_4$ control the balance between the three losses and are set to 0.15 with the same method as described previously.



\section{Experiments Results}
\subsection{Datasets and Evaluation Metrics}
\begin{figure}
\centerline{\includegraphics[width=0.98\columnwidth]{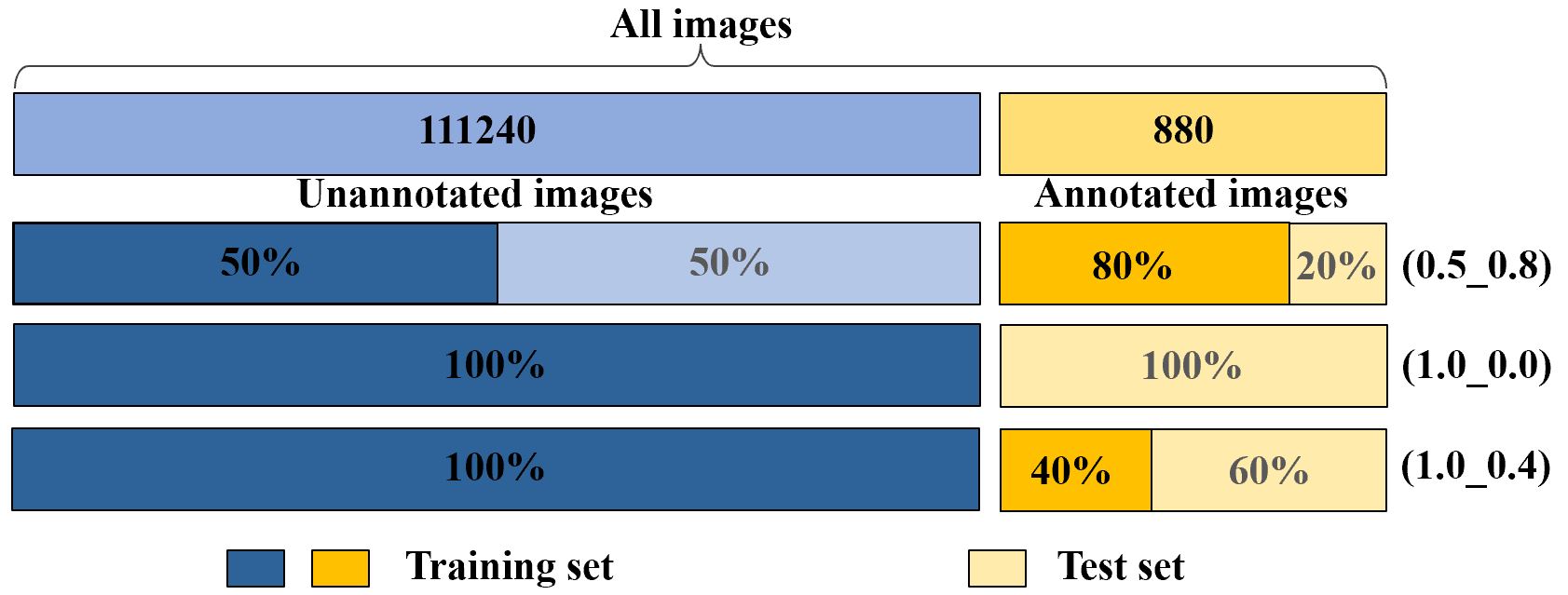}}
\caption{Three participation methods.}
\label{fig8}
\end{figure}

The NIH chest X-ray dataset~\cite{wang2017chestx} consists of 112,120 frontal-view X-ray images with 14 classes of diseases. Furthermore, the dataset contains 880 images with 984 labeled bounding boxes, and the provided bounding boxes only have 8 type of disease instances. Since we pay more attention to the task of locating diseases, we follow the terms in~\cite{li2018thoracic} and~\cite{liu2019align} to call the 880 images with labeled bounding boxes as ‘annotated' and the remaining 111,240 images as ‘unannotated' (Figure \ref{fig8}). For fast processing, we resize the original 3-channel images from resolution of $1024\times 1024$ to $512\times 512$ without any data augmentation techniques.

We follow the metrics used in~\cite{li2018thoracic}. For localization, the intersection over union (IoU) between predictions and ground truths is used to evaluate the performance of the models. The localization results are regarded as correct when $IoU > T(IoU)$, where $T(*)$ is the threshold. In practice, we usually pay most attention to the accuracy of high thresholds of T(IoU), so two thresholds of 0.5 and 0.7 are set in our experiments. For more accurate localization with limited data with location annotations, the localization predictions are discrete small rectangles. The performance of the eight diseases with ground truth boxes is reported, which is inherited from existing researches~\cite{li2018thoracic} and~\cite{liu2019align}.

\subsection{Experimental Settings}
All the models are trained on the NIH chest X-ray dataset using the stochastic gradient descent (SGD) algorithm with the Nesterov momentum. The learning rate starts from 0.001 and decreases by 10 times after every 4 epochs with a total of 9 epochs. The weight decay is 0.0001 and the momentum is 0.9. All the weights are initialized with pre-trained ResNet-50~\cite{he2016deep} on ImageNet~\cite{deng2009imagenet}. The mini-batch size is set to 4 with the NVIDIA 2080Ti GPU. All the algorithms are implemented with PyTorch~\cite{paszke2017automatic}. The threshold of 0.5 is used to distinguish positive grids from negative grids in the class-wise feature map, which has been adopted in previous studies~\cite{li2018thoracic} and~\cite{liu2019align}. Note that the up-sampling operation~\cite{liu2019align} of feature maps is used to gain a more accurate localization for lesion location before two last fully convolutional layers. The paired images are randomly sampled in the train phase, and the test phase does not need the paired images. 
\subsection{Comparison with State-of-the-Art}

In order to effectively evaluate GREN in weakly supervised disease localization, we strictly follow~\cite{li2018thoracic},~\cite{DBLP:conf/mm/Zhao21},~\cite{liu2019align} and use three data participation methods (Figure \ref{fig8}) and compute the average values with a \emph{five-fold cross validation} in each participation. In the first participation, we use 50\% unannotated images and 80\% annotated images for training and the remaining 20\% annotated images for testing (0.5\_0.8). In the second participation, we use the 100\% unannotated images without annotated images for training and all annotated images for testing (1.0\_0.0). In the third participation, we use the 100\% unannotated images and 40\% annotated images for training and the remaining 60\% annotated images for testing (1.0\_0.4).

Our experiment results are compared with six state-of-the-art methods. The first method is Wang et al.~\cite{wang2017chestx}, which uses a multi-label disease classification model with disease heatmaps for location disease. The second method is Li et al.~\cite{li2018thoracic}, which integrates the multi-instance level loss and binary cross-entropy loss into one framework for disease localization. The third method is Liu et al.~\cite{liu2019align}, which utilizes the contrast-induced attention acquired on paired images between healthy and unhealthy samples to provide information for localization. The fourth method is Zhou et al.~\cite{zhou2021contrast}, which exploits two contrastive abnormal attention models and a dual-weighting graph convolution to improve the performance of thoracic multi-disease recognition. The fifth method is Zhao et al.~\cite{zhao2021contralaterally}, which proposes the contralateral context information for enhancing feature representations of disease proposals by using a spatial transformer network. The sixth method is Zhao et al.~\cite{DBLP:conf/mm/Zhao21}, which proposed the CCG to regularize the model using the similarities computed via of clustered regions. 
\begin{table*}
\center
\caption{Performance comparison of disease localization using unannotated images only.}
\resizebox{0.98\textwidth}{!}{
\begin{tabular}{l c c c c c c c c c c}
\hline
T(IoU) & Models & Atelectasis& Cardiomegaly & Effusion & Infiltration & Mass & Nodule & Pneumonia & Pneumothorax & Mean \\
\hline
\multirow{8}{*}{ 0.5}& Wang et al.$^*$~\cite{wang2017chestx} & 0.00 & 0.25 & 0.01 & 0.06 & 0.09 & 0.05 & 0.06 & 0.00 & 0.07 \\
	& Li et al.$^*$~\cite{li2018thoracic}& 0.18& 0.10 & 0.27 & 0.46 & 0.18 & 0.03 & 0.00 & 0.11 & 0.17 \\
	& Liu et al.~\cite{liu2019align} & 0.19 & 0.53 & 0.19 & 0.47 & 0.33 & 0.03 & 0.08 & 0.11 & 0.24 \\
	& Zhao et al.~\cite{zhao2021contralaterally} & 0.32 & 0.68 & 0.39 & 0.61 & \textbf{0.49} & 0.07 & 0.15 & 0.21 & 0.36 \\ 
& Zhao et al.~\cite{DBLP:conf/mm/Zhao21} & 0.31 & 0.79 & 0.37 & \textbf{0.75} & 0.40 & 0.06 & 0.24 & \textbf{0.27} & 0.40\\
	\cline{2-11}  
	& Baseline (\textbf{B}) & 0.18 & 0.51 & 0.14 & 0.47 & 0.27 & 0.03 & 0.01 & 0.12 & 0.22\\ 
    & \textbf{B}+Intra & 0.30 & 0.79 & 0.38 & 0.72 & 0.42 & 0.05 &\textbf{0.25} & \textbf{0.27} & 0.40 \\
	& \textbf{B}+Inter & 0.31 & 0.79 & 0.37 & 0.70 & 0.41 & 0.08 & 0.08 & 0.24 & 0.37 \\
	& GREN\ (\textbf{B}+Intra+Inter) & \textbf{0.36} & \textbf{0.86} & \textbf{0.41} & 0.70 & \textbf{0.49} & \textbf{0.10} & 0.11 & 0.22 & {\color{red}\textbf{0.41}} \\
\hline
\multirow{8}{*}{ 0.7}& Wang et al.$^*$~\cite{wang2017chestx} & 0.00 & 0.01 & 0.00 & 0.03 & 0.04 & 0.04 & 0.01 & 0.00 & 0.01 \\
	& Li et al.$^*$~\cite{li2018thoracic} & 0.09& 0.01 & 0.07 & 0.28 & 0.08 & 0.01 & 0.00 & 0.05 & 0.07 \\
	& Liu et al.~\cite{liu2019align} & 0.08 & 0.30 & 0.09 & 0.25 & 0.19 & 0.01 & 0.04 & 0.07 & 0.13 \\  
	& Zhao et al.~\cite{zhao2021contralaterally} & 0.13 & 0.54 & 0.19 & 0.27 & 0.27 & 0.04 & 0.05 & 0.17 & 0.21 \\     
& Zhao et al.~\cite{DBLP:conf/mm/Zhao21} & 0.06 & 0.64 & 0.08 & 0.38 & 0.19 & 0.01 & 0.08 & 0.09 & 0.19\\
	\cline{2-11}  
	& Baseline (\textbf{B})& 0.11 & 0.34 & 0.06 & 0.32 & 0.20 & 0.01 & 0.00 & 0.06 & 0.14\\ 
    & \textbf{B}+Intra & 0.21 & 0.74 & 0.22 & \textbf{0.62} & 0.35 & 0.03 &\textbf{0.18} & 0.18 & 0.32 \\
	& \textbf{B}+Inter & 0.20 & 0.75 & \textbf{0.25} & 0.55 & 0.38 & 0.04 & 0.06 & \textbf{0.21} & 0.30 \\
	& GREN\ (\textbf{B}+Intra+Inter) & \textbf{0.28} & \textbf{0.77} & 0.22 & 0.61 & \textbf{0.44} & \textbf{0.06} & 0.09 & 0.16 & {\color{red}\textbf{0.33}} \\
\hline
\end{tabular}}
\label{table 2}
\end{table*}

Here, a model needs to predict the location of eight diseases, so we pay more attention to the overall performance of the model rather than to a particular disease. Therefore, in the following discussion, we are most concerned with the mean accuracy. In the first participation, we compare the localization results of our model with~\cite{wang2017chestx},~\cite{li2018thoracic},~\cite{liu2019align},~\cite{zhou2021contrast} and~\cite{DBLP:conf/mm/Zhao21}. We can observe that GREN outperforms the existing methods in most cases, as shown in Table \ref{table 1}. Particularly, with the increase of T(IoU), GREN has greater advantages over the reference models. For examples, when T(IoU) = 0.5, the mean accuracy is 0.54, and outperforms~\cite{wang2017chestx},~\cite{li2018thoracic},~\cite{liu2019align},~\cite{zhou2021contrast} and~\cite{DBLP:conf/mm/Zhao21} by 0.48, 0.27, 0.15, 0.09 and 0.05. 
However, when T(IoU) = 0.7, the mean accuracy of GREN is 0.46, and outperforms~\cite{wang2017chestx},~\cite{li2018thoracic},~\cite{liu2019align},~\cite{zhou2021contrast} and~\cite{DBLP:conf/mm/Zhao21} by 0.45, 0.34, 0.17, 0.11 and 0.02. In the second participation, we train our model without any annotated images. Since~\cite{li2018thoracic} only provides the results at T(IoU) = 0.1, in order to better show the performance of our model, we reproduce the results of T(IoU) = 0.5 and 0.7. Note that the sysbol $*$ denotes the reproduced results. For each disease, the bold and red values denote the best results of the eight disease types and mean accuracy, respectively. We compare the localization results of our model with~\cite{wang2017chestx},~\cite{li2018thoracic},~\cite{liu2019align},~\cite{zhao2021contralaterally} and~\cite{DBLP:conf/mm/Zhao21}. It can be seen that GREN outperforms existing methods in most cases, as shown in Table \ref{table 2}. 
In the third participation, we use more annotated images comparing the second participation. We compare the localization results of GREN with~\cite{li2018thoracic},~\cite{liu2019align} and~\cite{DBLP:conf/mm/Zhao21} in exact data setting. It can be seen that GREN still has the advantages over the existing methods, as shown in Table \ref{table 3}.

Overall, the experiment results demonstrate that GREN achieves state-of-the-art results for disease localization. Even when there is no annotated data for training, GREN still achieves decent localization results. Specifically, compared with the methods without image relations, the methods of modeling the relations in X-ray image have better performance. For example, the mean accuracies of~\cite{liu2019align},~\cite{DBLP:conf/mm/Zhao21},~\cite{zhao2021contralaterally},~\cite{zhou2021contrast} and GREN outperform~\cite{wang2017chestx},~\cite{li2018thoracic}. The comparative results show that image relations are useful in weakly-supervised medical image analysis. Compared with the methods that using only inter-image or intra-image relationship~\cite{liu2019align},~\cite{DBLP:conf/mm/Zhao21},~\cite{zhao2021contralaterally},~\cite{zhou2021contrast}, GREN that integrates the intra-image and inter-image relationship is obviously superior to these methods. It can be concluded that GREN is effectively applied to the weakly supervised disease localization tasks.

\begin{table*}
\center
\caption{Performance comparison of disease localization using 100\% unannotated images and 40\% annotated images. For each column, the bold or red values denote the best results. Note that the sysbol $*$ denotes the reproduced results.}
\resizebox{0.98\textwidth}{!}{
\begin{tabular}{l c c c c c c c c c c}
\hline
T(IoU) & Models & Atelectasis& Cardiomegaly & Effusion & Infiltration & Mass & Nodule & Pneumonia & Pneumothorax & Mean \\
\hline
\multirow{6}{*}{ 0.5}& Li et al.$^*$~\cite{li2018thoracic} & 0.23 & 0.72 & 0.30 & 0.60 & 0.22 & 0.02 & 0.32 & 0.20 & 0.32\\
	& Liu et al.~\cite{liu2019align} & 0.36 & 0.57 & 0.37 & 0.62 & 0.34 & 0.13 & 0.23 & 0.17 & 0.35 \\	       
	& Zhao et al.~\cite{DBLP:conf/mm/Zhao21} & 0.26 & 0.80 & 0.41 & 0.67 & 0.15 & 0.06 & 0.42 & 0.18 & 0.37 \\
 \cline{2-11}        
	& Baseline (\textbf{B})  & 0.27 & 0.76 & 0.39 & 0.58 & 0.24 & 0.02 & 0.39 & 0.21 & 0.36\\
    & \textbf{B}+Intra & 0.44 & \textbf{0.91} & 0.60 & 0.82 & 0.52 & 0.13 & 0.58 & 0.33 & 0.54 \\
	& \textbf{B}+Inter & \textbf{0.45} & 0.74 & \textbf{0.62} & 0.85 & 0.43 & \textbf{0.19} & 0.70 & 0.33 & 0.54 \\
	& GREN\ (\textbf{B}+Intra+Inter) & 0.42 & 0.83 & 0.60 & \textbf{0.86} & \textbf{0.54} & 0.10 & \textbf{0.71} & \textbf{0.36} & {\color{red}\textbf{0.55}} \\
\hline
\multirow{6}{*}{ 0.7}& Li et al.$^*$~\cite{li2018thoracic} & 0.07 & 0.64 & 0.17 & 0.38 & 0.17 & 0.00 & 0.20 & 0.17 & 0.21\\
   	& Liu et al.~\cite{liu2019align} & 0.19 & 0.47 & 0.20 & 0.41 & 0.22 & 0.06 & 0.12 & 0.11 & 0.22 \\ 
	& Zhao et al.~\cite{DBLP:conf/mm/Zhao21}  & 0.18 & 0.71 & 0.20 & 0.50 & 0.20 & 0.02 & 0.29 & 0.06 & 0.27\\
   	\cline{2-11}  
	& Baseline (\textbf{B})  & 0.14 & 0.62 & 0.20 & 0.42 & 0.07 & 0.00 & 0.23 & 0.08 & 0.22 \\ 
	& \textbf{B}+Intra & 0.35 & \textbf{0.89} & 0.49 & \textbf{0.76} & 0.37 & 0.06 & 0.51 & \textbf{0.26} & 0.46\\
	& \textbf{B}+Inter & \textbf{0.38} & 0.71 & 0.49 & 0.72 & \textbf{0.43} & \textbf{0.13} & 0.58 & 0.24 & 0.46 \\
	& GREN\ (\textbf{B}+Intra+Inter) & 0.32 & 0.78 & \textbf{0.52} & 0.74 & \textbf{0.43} & 0.06 & \textbf{0.65} & \textbf{0.26} & {\color{red}\textbf{0.47}} \\
\hline
\end{tabular}}
\label{table 3}
\end{table*}


\subsection{Ablation Studies}
\subsubsection{Ablation of the Inter and Intra Construction} 
We compare the localization results of the baseline model (B) with three models under different configurations, including the model with the intra-image knowledge learning module (B+Intra), the model with the inter-image knowledge learning module (B+Inter), and the model with both knowledge learning modules (B+Intra+Inter). The experiment results in Table \ref{table 1}, Table \ref{table 2} and Table \ref{table 3} show that the model (B+Intra) and the model (B+Inter) demonstrate more advantages over the baseline method, and the model (B+Intra+Inter) achieves the best localization result. For example, in Table \ref{table 1}, when T(IoU) = 0.5, the mean accuracy of the model (B+Intra) and the model (B+Inter) are 0.49 and 0.50, and outperform the baseline (B) by 0.21 and 0.22, respectively. The model (B+Intra+Inter) achieves the best localization results, which is 0.54, and outperforms the baseline (B) by 0.26. Additionally, the localization results in Table \ref{table 1} and Table \ref{table 2} demonstrate that when there is more annotated data in the training process, the model using the inter-image knowledge learning module has more advantages; when there is no annotated data in the training process, the model using the intra-image knowledge learning module has more advantages. For example, the model (B+Inter) performs better in most cases compared with the model (B+Intra) in Table \ref{table 1}, and the model (B+Intra) performs better in most cases compared with the model (B+Inter) in Table \ref{table 2}. Overall, the experiment results demonstrate that using intra-image and inter-image structural relationship information can improve the performance of models for disease localization. Moreover, the experiment results show that the combined effect of the inter and intra construction can further boost the performance of weakly supervised disease localization.


\begin{figure*}
\centerline{\includegraphics[width=1.0\textwidth]{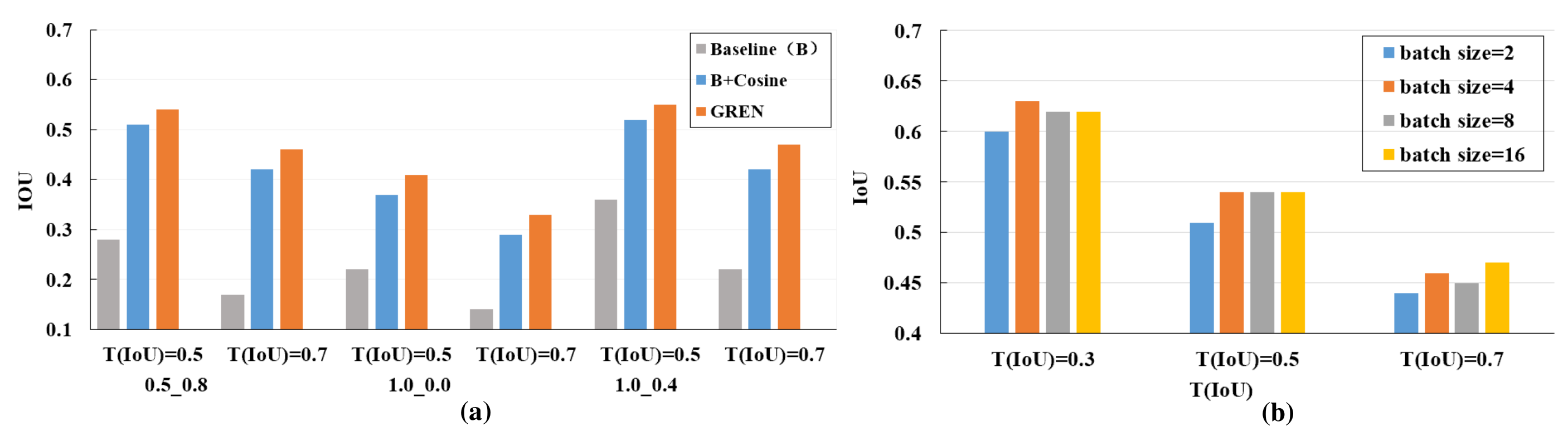}}
\caption{(a) The performance comparison of the baseline model, the CNN model (B+Cosine) and GREN in three data settings, each of which includes the mean accuracies with T(IOU)=0.5 and 0.7. (b) The performance comparison of different batch sizes. The batch size is set to 2, 4, 8, and 16. The mean accuracies improve with the increase of batch size, but the trend is not an unbounded growth. Note that the data partition settings are inherited from existing studies for fair comparisons.}
\label{fig3_2}
\end{figure*}

\begin{table*}[!t]
\center
\caption{Performance comparison of disease localization using 50\% unannotated images and 80\% annotated images.}
\resizebox{0.98\textwidth}{!}{
\begin{tabular}{l c c c c c c c c c c}
\hline
T (IoU) & Models & Atelectasis& Cardiomegaly & Effusion & Infiltration & Mass & Nodule & Pneumonia & Pneumothorax & Mean \\
\hline
\multirow{3}{*}{ 0.5} & Baseline (\textbf{B}) & 0.12 & 0.79 & 0.32 & 0.36 & 0.26 & 0.07 & 0.13 & 0.20 & 0.28 \\
	& \textbf{B}+Attention & 0.34 & \textbf{0.89} & \textbf{0.52} & 0.72 & 0.37 & 0.21 & \textbf{0.71} & 0.15 & 0.49 \\
	& GREN  &\textbf{0.37} & 0.86 & \textbf{0.52} & \textbf{0.84} & \textbf{0.42} & \textbf{0.29} & 0.54 & \textbf{0.45} & {\color{red}\textbf{0.54}}  \\
	
\hline
\multirow{3}{*}{ 0.7} & Baseline (\textbf{B}) & 0.05 & 0.71 & 0.16 & 0.04 & 0.16 & 0.00 & 0.08 & 0.15 & 0.17 \\
	& \textbf{B}+Attention & 0.29 & \textbf{0.89} & 0.44 & \textbf{0.64} & 0.32 & 0.14 & \textbf{0.58} & 0.15 & 0.43\\
	& GREN & \textbf{0.34} & 0.86 & \textbf{0.48} & 0.60 & \textbf{0.37} & \textbf{0.21} & 0.46 & \textbf{0.35} & {\color{red}\textbf{0.46}} \\
	
\hline
\end{tabular}}
\label{table 4}
\end{table*}
\subsubsection{Ablation of Similarity Computing Methods}
To investigate the similarity computing method, we compare the Hash coding using Hamming distance with CNN features using cosine similarity (B+Cosine), which is shown in Figure \ref{fig3_2}(a). The B+Cosine model is competitive to the baseline model but is inferior to GREN for all the three data settings. There are two possible reasons. First, it is not easy to obtain a CNN-based model that needs to not only perform as a feature extractor but also learn class-aware similarity. Hash coding solved this problem via projecting features to a low frequency signature. Second, as a comparison, Hash coding is data-independent and training-free. Overall, the structural relationship (similarity) can be used to regularize the training of deep neural networks, so as to achieve more accurate and integral localization results with few careful annotations. Moreover, using Hash coding and Hamming distance instead of the cosine distance of CNN features can achieve better performance, which demonstrates that the image features extracted directly using Hash coding can to some extent compensate for the information lost in CNNs, making the proposed method reasonable. 

\subsubsection{The Number of Graph Nodes}
We explore the influence of different number of graph nodes $N$ on the results, which is shown in Figure \ref{fig3_2}(b). The value is equal to the batch size of the model, including 2, 4, 8, and 16. The data of 100\% unannotated images and 40\% annotated images are used to evaluate the performance. The mean accuracies improve with the increase of batch size, whereas the trend is not an unbounded growth. For example, when T(IoU) = 0.7, the mean accuracy of the model with batch size of 16 is 0.47, outperforming the model with batch size of 2 by 0.03. However, when T(IoU) = 0.5, the mean accuracy of the model with batch size of 16 is 0.54, which is the same as the model with the batch size of 4. Using more images in each graph is good to effective regularization, but the growth is bounded (slow after the batch size exceeds 4). Therefore, the batch size is set to 4 in our experiment by considering the trade-off between memory cost burden and performance.

\subsubsection{Ablation of Inter-image and Intra-image Similarity}
Graph is a concise and effective way to characterize inter- and intra-image relations. There are several works to investigate the relationship and achieve a superior performance in multi-label natural image recognition, such as~\cite{Semantic-Interactive9524695}. In the X-ray image analysis, the inter- and intra-image relations can be used to simulate the physician's practice of comparing multiple images and symmetric regions of the same image for diagnosis respectively. Existing literatures~\cite{zhao2021contralaterally},~\cite{DBLP:conf/mm/Zhao21} imply that the relationship between the left and right lung lobes is useful in automatic diagnosis. However, these methods do not explore the underlying reasons why the inter- and intra- relationship modeling is effective. Here, we construct relation graphs using the left and right lung lobes as the nodes of graphs, and then incorporate the graph as a regularizer to facilitate the representation learning. Specially, we try to explore the underlying reasons of this. 

\begin{figure}[!t]
\centerline{\includegraphics[width=0.95\columnwidth]{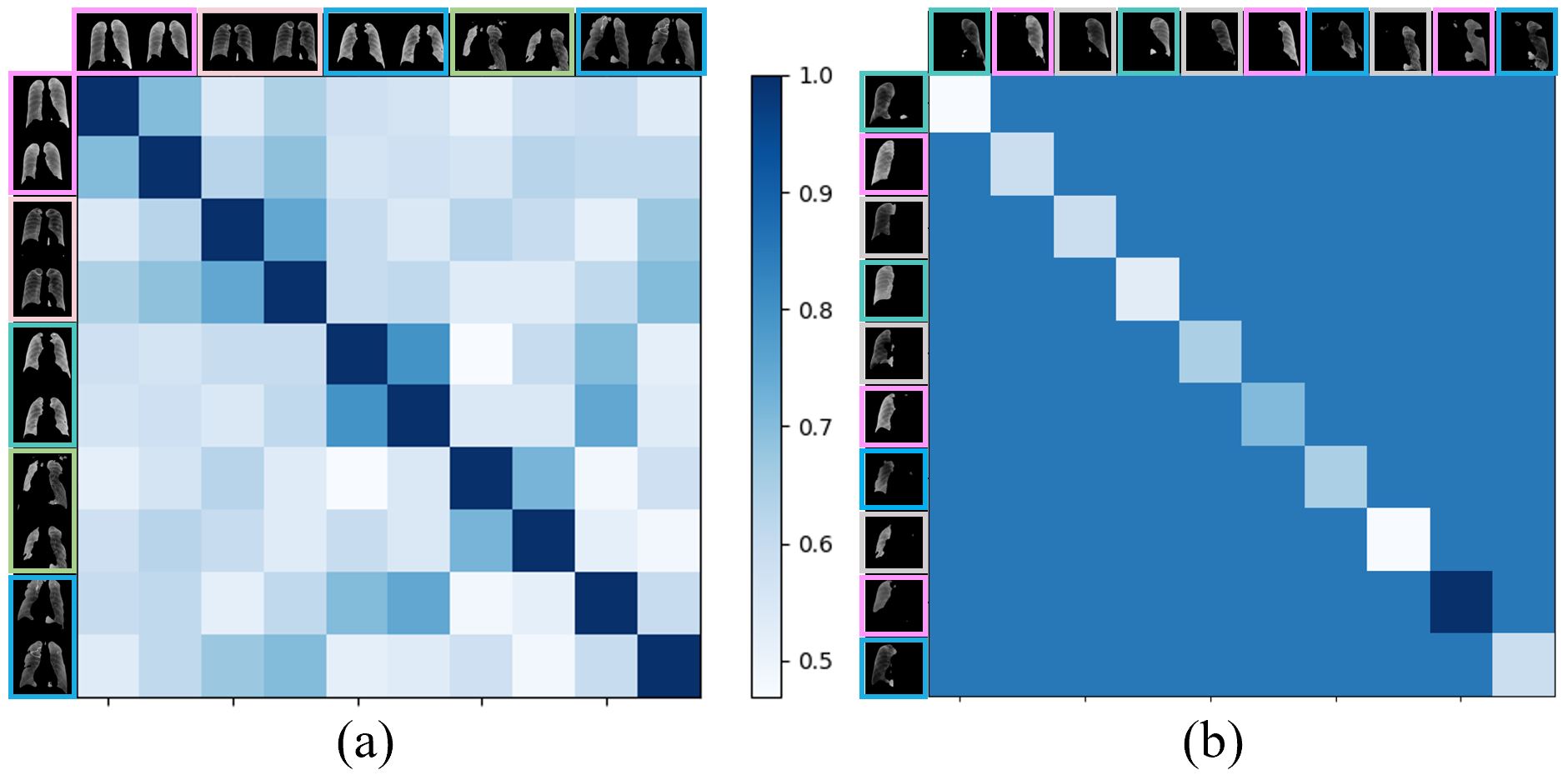}}
\caption{The inter-image and intra-image similarities. Each color denotes a value of similarity, where the darker color denotes higher similarity. (a). The inter-image similarity. The horizontal and longitudinal axis are the same lung region images. The two adjacent images have the same classes and higher similarity compared with other images. (b). The intra-image similarity. The horizontal and longitudinal axis are the left lung region and right lung region in an image. Please note that the same color boxes denote the same classes of the images and the grey boxes denote the image only has one class.} 
\label{fig4}
\end{figure}

We explore the influence of the inter-image and intra-image similarity as shown in Figure \ref{fig4}. We randomly choose 10 images from the dataset. In Figure \ref{fig4}(a), the horizontal and longitudinal axis are the same lung region images, and the two adjacent images have the same classes. We calculate the similarity of lung regions of these images using the Hash coding and Hamming distance. The two adjacent images have higher similarity compared with other images, where the darker color denotes higher similarity. For example, the diagonal rectangles are the similarities of the same images, so the values are 1. Additionally, we calculate the similarity of the left lung and right lung regions in an image, which is diagonal rectangle, and other rectangles are filled with the same color. In Figure \ref{fig4}(b), the horizontal and longitudinal axis are the corresponding left lung and right lung regions in an image. The darker-colored rectangle denotes higher similarity of the left lung and right lung in an image. Particularly, the sixth and ninth darker color rectangles (from upper left to low right) imply that the sixth and ninth images have highly similar left and right lung, and their categories are actually “No Finding" (The categories of the second, sixth, and ninth images are “No Finding"). The first, fourth, and eighth light color rectangles imply that the left and right lung regions of the first, fourth, and eighth images have low similarity, and their categories are “Cardiomegaly", “Cardiomegaly", and “Pneumonia", respectively. To some degree, the inter-image similarity can indicate the categories of images without annotation data, and the intra-image similarity can indicate the information of differences of left and right lung regions, which is helpful to distinguish the unilateral disease of X-ray images.

\begin{table*}[!t]
\center
\caption{The AUC (\%) scores of the baseline model and GREN. The bold or red values denote the best results.}
\resizebox{0.98\textwidth}{!}{

\begin{tabular}{ c c c c c c c c c}
\hline
 Models & Atelectasis & Cardiomegaly & Consolidation & Edema & Effusion & Emphysema & Fibrosis & \\  
\hline
 
	 Baseline (\textbf{B})  & 76.77 & 87.09 & 77.73 & 87.90 & 84.65 & 90.04 & 77.87 & \\
	 \textbf{B}+Intra   & 77.39 & 87.78 & 76.16 & 87.13 & \textbf{85.24} & 90.98 & 77.83 & \\
	 \textbf{B}+Inter   & 76.61 & 86.17 & 77.02 & 89.00 & 84.16 & 90.22 & 79.43 & \\
	 GREN & \textbf{77.57} & \textbf{88.19} & \textbf{79.01} & \textbf{89.25} & 85.11 & \textbf{91.98} & \textbf{79.50} &    \\
\hline
 Models & Hernia & Infiltration & Mass & Nodule & Pleural Thickening & Pneumonia & Pneumothorax & Mean\\
\hline
	 Baseline (\textbf{B})   & 88.71 & 64.53 & 80.57 & 68.37 & 77.68 & 67.62 & 85.62 & 79.65\\ 
	 \textbf{B}+Intra    & \textbf{91.84} & 64.84 & 80.59 & \textbf{73.01} & 77.62 & 67.06 & 85.99 & 80.25\\ 
	 \textbf{B}+Inter    & 81.15 & 64.08 & 81.23 & 70.63 & 78.53 & \textbf{72.36} & 86.40 & 80.32\\ 
	 GREN  & 86.65 & \textbf{66.04} & \textbf{81.34} & 72.07 & \textbf{78.96} & 67.47 & \textbf{87.25} & {\color{red}\textbf{80.74}} \\
\hline
\end{tabular}}
\label{table 5}
\end{table*}

\begin{figure*}[!t]
\centerline{\includegraphics[width=0.95\textwidth]{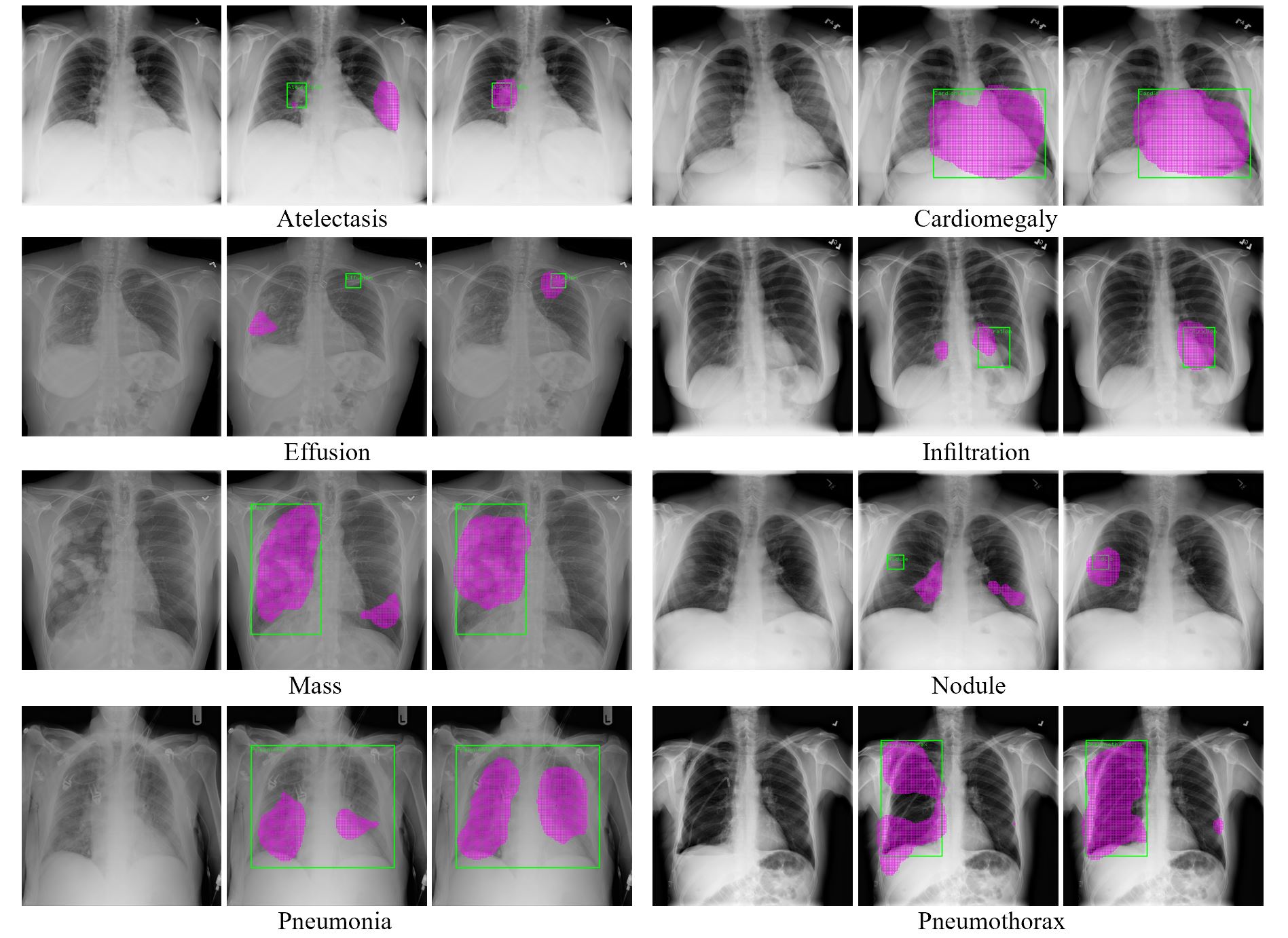}}
\caption{Visualizations of the predicted results on both the baseline model and our method. The first column shows the original images, the second and third columns show the predicted results of the Baseline and GREN. The green bounding box and red area mean the ground truth and prediction, respectively. }
\label{fig5}
\end{figure*}

\subsubsection{Ablation of Attention-based Method}
We compare the localization results of our model with the attention-based method, which integrates the attention mechanism into the baseline to learn the difference of lung regions. In Table \ref{table 4}, the B+Attention model outperforms the baseline by 0.21 and 0.26 at T(IoU) = 0.5 and 0.7, however, GREN still holds advantages. It ourperforms the B+Attention model by 0.26 and 0.29 under the two thresholds. Overall, the attention-based method can indeed improve the accuracy, but GREN using the intra-image and inter-image structural relationship has larger advantages to enhance the weakly-supervised disease localization. 

\begin{figure*}[!t]
\centerline{\includegraphics[width=0.95\textwidth]{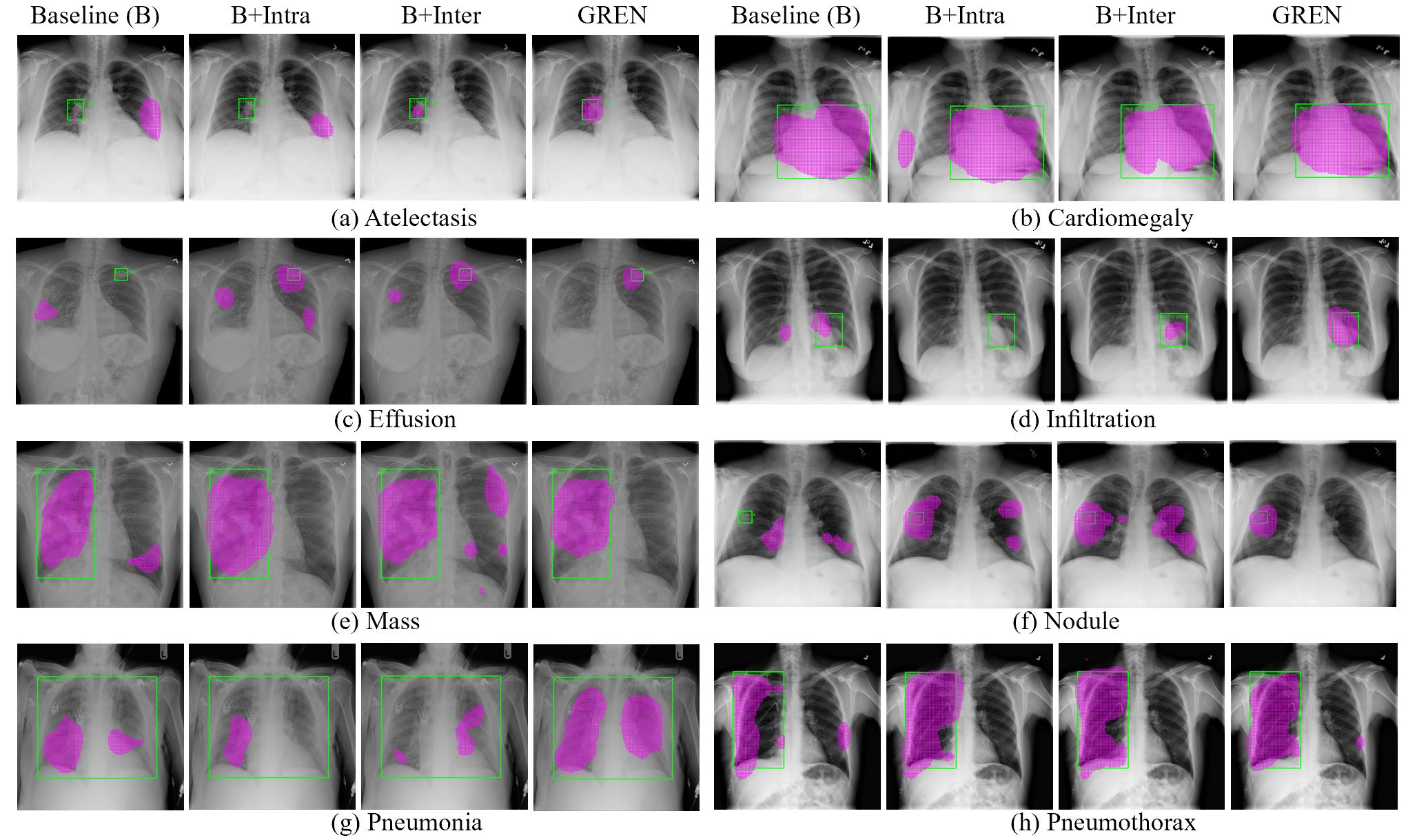}}
\caption{Visualizations of the predicted results on Baseline (B), B+Inter, B+Intra and GREN. From left to right, it shows the localization results of the baseline model, B+Intra, B+Inter and GREN. The green bounding box and red area mean the ground truth and prediction, respectively.} 
\label{fig10}
\end{figure*}

\begin{figure*}[!t]
\centerline{\includegraphics[width=0.95\textwidth]{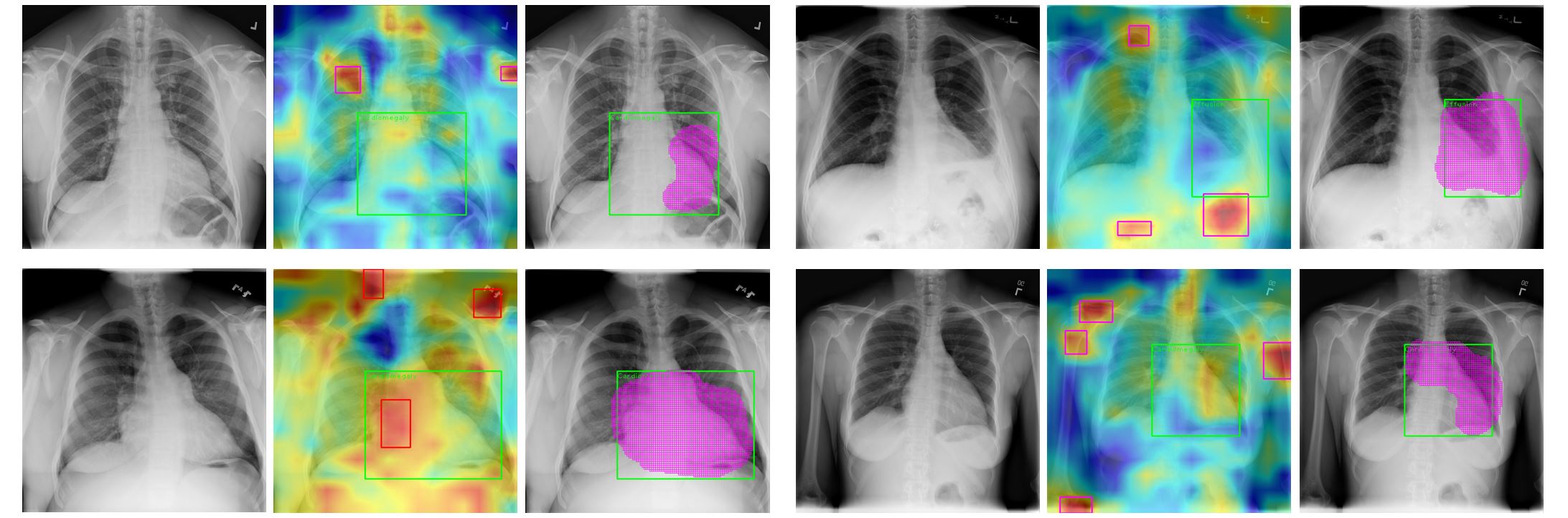}}
\caption{visualizations of the predicted results on both~\cite{wang2017chestx} and GREN. The first column shows the original images, the second and third columns show the predicted results of the~\cite{wang2017chestx} and GREN. The green bounding box and red area mean the ground truth and prediction, respectively. ~\cite{wang2017chestx} only highlights the most discriminative areas using CAM, and GREN makes more integral predictions.}
\label{fig6}
\end{figure*}

\subsubsection{Disease Identification Results}
Although we focus on the weakly-supervised localization task, we additionally evaluate the results of disease identification. Table \ref{table 5} presents the AUC scores (the area under the receiver operating characteristic curve) for all the classes. Following~\cite{li2018thoracic} and~\cite{liu2019align}, we use 70\% images for training and 20\% images for testing. We compare the performance of disease identification of the Baseline (B) with the B+Intra, B+Inter and GREN. The mean AUC scores of the B+Intra, B+Inter and GREN are 80.25\%, 80.32\% and 80.74\%, outperforming the Baseline (79.65\%) by 0.60\%, 0.67\% and 1.09\%, respectively. GREN is more effective in disease localization than disease identification because the task of disease identification has more sufficient supervision. Overall, GREN can maintain performance of disease identification, but enjoys better ability for weakly-supervised disease localization. Particularly, GREN has better application value in the task with limited annotation data. 
\subsubsection{Visualizations}
To better demonstrate the final effect of our method on disease localization, we visualize some of typical predictions of both the baseline model and GREN, as shown in Figure \ref{fig5}. The first column shows the original images, the second and the third column show the localization results of the baseline model and GREN. The green bounding box and red area mean the ground truth and prediction, respectively. It can be seen that GREN can predict more accurate regions compared with the baseline model. 
For example, for the class of “Nodule", the localization results of the baseline model are completely inconsistent with the ground truth, but the localization results of GREN are consistent with the ground truth. 

In order to demonstrate the difference of GREN (B+Intra+Inter) with B+Intra and B+Inter, we visualize some of typical predictions of them, as shown in Figure \ref{fig10}. From left to right, it shows the localization results of the Baseline (B), B+Intra, B+Inter and GREN. GREN plays an important role in integrating the advantages of B+Inter and B+Intra to make the predictions better. For a sample (Figure \ref{fig10}), some of B+Inter results are better than B+Intra results, such as (a), (b), (d), some of B+Intra results are better than B+Inter results, such as (e), (f). It can be seen that GREN can modify and complement the predicted results of B+Inter and B+Intra to make better predictions. The underlying reason may be that the functions of Inter-image and Intra-image knowledge learning module are not only complementary but also share common mechanisms. For example, the lungs are highly structured. For an X-ray image, there are not only similarities but also differences between left and right lungs. For different X-ray images, there are not only differences but also similarities in ipsilateral lung. Therefore, the functions of Inter-image and Intra-image knowledge learning module cannot be completely separated, and some overlap and crossover inevitably exist between them.

We also visualize some of typical predictions of both the CAM-based method~\cite{wang2017chestx} and GREN, as shown in Figure \ref{fig6}. The first column shows the original images, the second and the third column show the localization results of~\cite{wang2017chestx} and GREN. It can be seen that GREN has greater advantages over~\cite{wang2017chestx}. Moreover, compared with~\cite{wang2017chestx}, GREN uses the regularization of the intra-image and inter-image structural information to pay more attention to the lung regions than other regions (e.g., clavicle and arm regions). It further demonstrates that GREN can effectively improve the performance of weakly supervised disease localization.

\section{Conclusion}
We have proposed and evaluated GREN, which leverages the intra-image and inter-image information to regularize CNNs to preserve the structural similarities between lung regions and image pairs in the embedding space. The experiment results on the NIH Chest-14 dataset demonstrate that GREN achieves the state-of-the-art performance in various settings. We demonstrate that the relationship of intra-image and inter-image can be used to model the graphs to facilitate training in weakly supervised disease localization. The proposed method has a very practical use in the weakly-supervised localization task. Since the public dataset available for X-ray lung segmentation only allowed us to segment the left and right lungs, we build graphs based on two lungs. However, many areas of the lung have explicit anatomical meanings, such as right upper lobe, right middle lobe, right lower lobe, left upper lobe, and left lower lobe. In theory, the more comprehensive lung regions are considered, the more relationship information GREN can model. For future work, we will exploit the relationship information of five lung lobes from CT images to construct graphs to compensate for the lack of labels in weakly supervised disease localization. We will also investigate algorithms applicable to the localization of small targets (especially pulmonary nodules).

{\small
\bibliographystyle{IEEEtranS}
\bibliography{egbib}
}

\end{document}